\documentclass[dvipsnames]{article} %
\usepackage{colm2024_conference}

\usepackage{booktabs}
\usepackage{enumitem}
\usepackage{wrapfig}
\usepackage{algorithm}
\usepackage{algpseudocode}
\usepackage{graphicx}
\usepackage{subfigure}
\usepackage[misc]{ifsym}
\usepackage{pdfpages}
\usepackage{fancyvrb}

\usepackage{listings}

\usepackage{microtype}
\usepackage{amsmath}
\usepackage{amssymb}
\usepackage{stmaryrd}
\usepackage{colortbl}
\usepackage[utf8]{inputenc}
\usepackage[T1]{fontenc}
\definecolor{lightgray}{rgb}{0.9,0.9,0.9}
\usepackage{caption}
\usepackage{subcaption}
\usepackage{setspace}
\usepackage{url}
\usepackage{multirow}
\usepackage{tabularx}
\usepackage{blindtext}
\usepackage{pgfplots}
\pgfplotsset{compat=1.18} 
\usepackage{tikz}
\usetikzlibrary{er,positioning,bayesnet}
\usepackage{makecell}
\usepackage{tipa}
\usepackage{siunitx}
\usepackage{nicefrac}
\usepackage{listings}
\usepackage[raster,skins, most]{tcolorbox} %
\usepackage{xltabular}
\usepackage{adjustbox}
\usepackage{xurl}
\usepackage{rotating}
\usepackage[normalem]{ulem}
\usepackage{booktabs}
\usepackage{fontawesome}

\useunder{\uline}{\ul}{}

\usepackage{amsmath,amsfonts,bm}

\def\eqref#1{equation~\ref{#1}}

\def\1{\bm{1}}

\DeclareMathAlphabet{\mathsfit}{\encodingdefault}{\sfdefault}{m}{sl}
\SetMathAlphabet{\mathsfit}{bold}{\encodingdefault}{\sfdefault}{bx}{n}

\newcommand*\justify{%
  \fontdimen2\font=0.4em
  \fontdimen3\font=0.2em
  \fontdimen4\font=0.1em
  \fontdimen7\font=0.1em
  \hyphenchar\font=`\-
}

\renewcommand{\texttt}[1]{%
  \begingroup
  \ttfamily
  \begingroup\lccode`~=`/\lowercase{\endgroup\def~}{/\discretionary{}{}{}}%
  \begingroup\lccode`~=`[\lowercase{\endgroup\def~}{[\discretionary{}{}{}}%
  \begingroup\lccode`~=`.\lowercase{\endgroup\def~}{.\discretionary{}{}{}}%
  \catcode`/=\active\catcode`[=\active\catcode`.=\active
  \justify\scantokens{#1\noexpand}%
  \endgroup
}

\usepackage{makecell}
\usetikzlibrary{tikzmark}
\makeatletter
\newcommand*\myfontsize{%
  \@setfontsize\myfontsize{7}{8}%
}
\makeatother

\definecolor{uclablue}{RGB}{159, 195, 224}

\definecolor{uclagold}{RGB}{255, 240, 180}

\definecolor{aliceblue}{RGB}{255, 238, 241}

\definecolor{cadmiumgreen}{rgb}{0.0, 0.42, 0.24}

\definecolor{myred}{rgb}{0.7, 0.3, 0.0}
\definecolor{myblue}{rgb}{0.2, 0.3, 0.6}
\definecolor{babygreen}{rgb}{0.85, 0.97, 0.85}

\definecolor{purple1}{RGB}{126, 107, 196}
\definecolor{purple2}{RGB}{199, 158, 207}
\definecolor{purple3}{RGB}{214, 200, 255}
\definecolor{purple4}{RGB}{254, 240, 255}

\definecolor{deepblue}{RGB}{48, 58, 82}

\hypersetup{ breaklinks, colorlinks=true, allcolors=uclablue_old }

\definecolor{deepPurple}{HTML}{330066}

\definecolor{uclablue_old}{rgb}{0.329, 0.318, 0.961}
\hypersetup{
    breaklinks,
    citecolor=uclablue_old,
    colorlinks=true,
}

\newtcolorbox{mybox}[2][]
  {colback = black!5!white, colframe = black!75!black, fonttitle = \bfseries,
    colbacktitle = black!100!black, enhanced, before upper={\fontsize{8}{11}\obeyspaces\obeylines\selectfont}, fontupper=\selectfont,
    attach boxed title to top left={yshift=-2.2mm,xshift=4mm},
    title=#2,#1}

\title{%
\raisebox{-1.2em}{
  \parbox[t]{0.3in}{\includegraphics[width=0.51in]{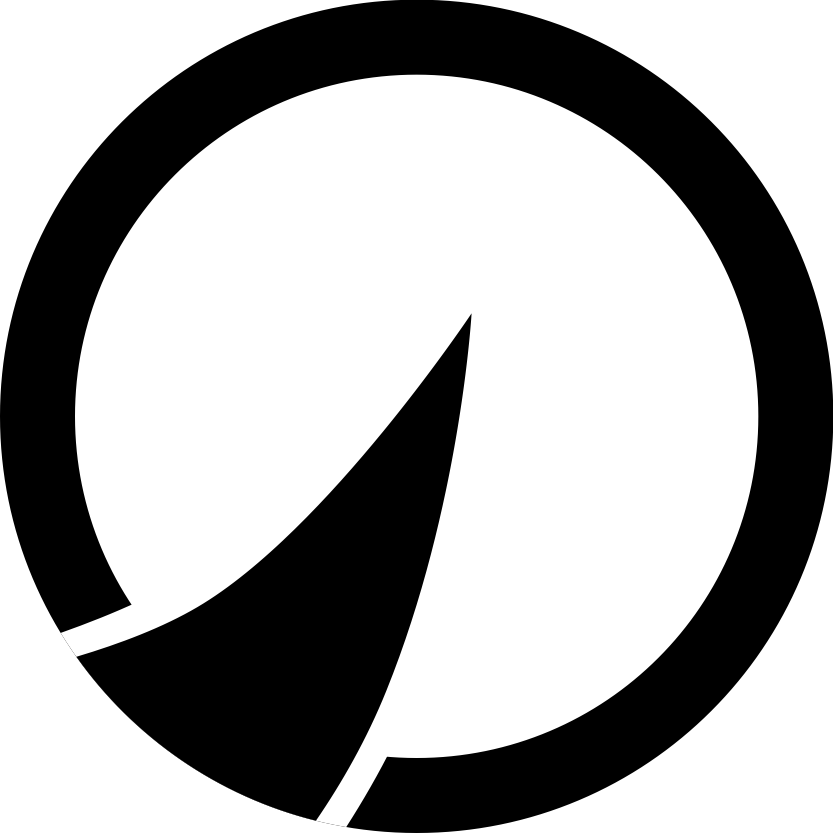}}\quad 
  }
  \begin{tabular}[t]{l}
  \parbox[t]{0.8\textwidth}{ 
    RhinoInsight: Improving Deep Research through \\ Control Mechanisms for Model Behavior and Context
  }
  \end{tabular}
}

\author{
\textbf{Yu Lei\textsuperscript{*}$^{\heartsuit}$$^\diamondsuit$, Shuzheng Si\textsuperscript{*}$^{\spadesuit}$$^{\heartsuit}$, Wei Wang\textsuperscript{*}$^{\heartsuit}$$^{\bigstar}$, Yifei Wu$^{\heartsuit}$, Gang Chen$^{\heartsuit}$ \\ Fanchao Qi$^{\heartsuit}$$^{\spadesuit}$$^{(\textrm{\Letter})}$, Maosong Sun$^{\spadesuit}$}
\\[0.5em]
{\fontsize{10pt}{11pt}\selectfont
$^{\heartsuit}$ DeepLang AI \\
$^{\spadesuit}$ Department of Computer Science and Technology, Tsinghua University \\ $^\diamondsuit$ Beijing University of Posts and Telecommunications $^{\bigstar}$ Beijing Jiaotong University 
}
}

\begin{document}
\vspace*{-0.8cm}

\maketitle

\begingroup
  \renewcommand\thefootnote{*}  
  \footnotetext{~Equal contribution.}
\endgroup

\begingroup
  \renewcommand\thefootnote{\Letter}  
  \footnotetext{~Corresponding author. Emails: fanchao.qi@deeplang.ai}
\endgroup

\vspace{-0.1in}
\begin{abstract}

Large language models are evolving from single-turn responders into tool-using agents capable of sustained reasoning and decision-making for deep research. Prevailing systems adopt a linear pipeline of plan to search to write to a report, which suffers from error accumulation and context rot due to the lack of explicit control over both model behavior and context. We introduce RhinoInsight, a deep research framework that adds two control mechanisms to enhance robustness, traceability, and overall quality without parameter updates. First, a Verifiable Checklist module transforms user requirements into traceable and verifiable sub-goals, incorporates human or LLM critics for refinement, and compiles a hierarchical outline to anchor subsequent actions and prevent non-executable planning. 
Second, an Evidence Audit module structures search content, iteratively updates the outline, and prunes noisy context, while a critic ranks and binds high-quality evidence to drafted content to ensure verifiability and reduce hallucinations. 
Our experiments demonstrate that RhinoInsight achieves state-of-the-art performance on deep research tasks while remaining competitive on deep search tasks.
\end{abstract}

\vspace{-0.8em}
\begingroup
\setlength{\textfloatsep}{8pt plus 2pt minus 2pt}%
\setlength{\intextsep}{8pt plus 2pt minus 2pt}%
\begin{figure}[H]
  \centering
  \includegraphics[width=\linewidth]{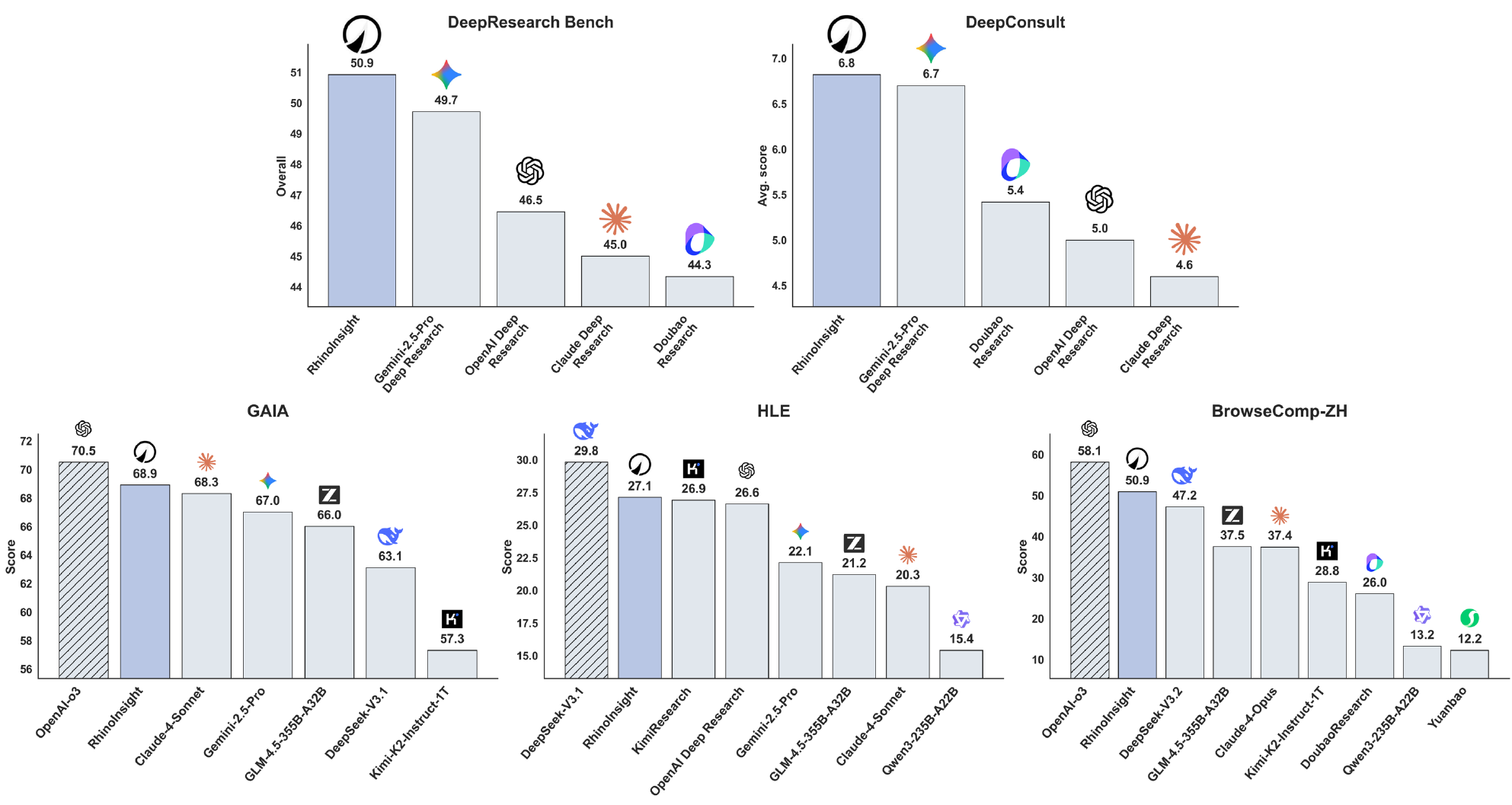}
  \caption{\textbf{RhinoInsight} shows the state-of-the-art performance in deep research tasks compared with proprietary systems, while remaining competitive on deep search tasks.}
  \label{fig:res_fig}
\end{figure}
\endgroup
\vspace{-0.8em}

\section{Introduction}
Large language models (LLMs) ~\citep{gpt5, deepseekv3.1, gemini_deep_research} are evolving from fluent generators into tool-using problem-solvers capable of sustained reasoning and decision-making. 
{In this work, we focus on deep research: an agentic capability that performs multi-step reasoning and searches for information on the internet to tackle complex research tasks.}
Deep research agents present an innovative approach to enhancing and potentially transforming human intellectual productivity. 

Prevailing approaches~\citep{chai2025scimaster, liu2025prosocial, DeepConsult,  xbench,lei2024fairmindsim, chai2025scimaster} mainly focus on using a linear pipeline with different actions, such as planning, retrieval, writing, and reporting.
As a result, these methods rely on powerful agent models to carry out each step in a serial manner, as shown in Figure \ref{fig:intro_fra} (a), in order to mitigate common issues in linear pipeline structures, such as error accumulation and context rot.
\textbf{In contrast to these approaches, we argue that the key to improving the effectiveness of the deep research framework does not lie in simply enhancing model capability, but rather in how to properly incorporate control mechanisms for both the context and model behavior.}
Specifically, different components may face distinct challenges in the absence of proper guidance.
(1) Planning lacks executable and checkable anchors, which makes generated goals unclear and easily leads to incorrect actions in subsequent model generation.
(2) Search, memory, and drifting modules often generate excessively long contexts that introduce substantial noisy or irrelevant content. 
As the context grows, the model struggles to maintain effectiveness due to context rot and performs poorly when processing lengthy and disorganized contexts.
(3) The writing module operates at the end of the pipeline and is therefore highly sensitive to the cumulative context produced by all preceding modules—including erroneous actions and the noisy content they generate. 
Without proper intervention, this accumulated noise may propagate into the final information aggregation stage, degrading the final report quality and undermining the reliability of the produced analysis.
Therefore, it is necessary to introduce control mechanisms to guide model behavior, e.g., actions, and efficiently organize contextual information, ensuring the entire process remains stable and effective.

In this paper, we introduce \textbf{RhinoInsight}, a deep research framework that involves two control mechanisms to drive overall improvement.
Specifically, we introduce a verifiable checklist module to supervise and control model behavior, and an evidence audit module to organize the context information across planning, retrieval, memory, outlining, and writing modules.
Within the proposed verifiable checklist module, we use a checklist generator to produce traceable and verifiable sub-goals, then further incorporate manual efforts (e.g., scenarios where user queries are unclear in real-world applications) or automated LLMs as a critic to check these sub-goals, and then the planning module compiles them into a hierarchical outline.
Unlike previous works that utilize a planning module to generate outlines to guide the following steps, our proposed approach can help ensure that each sub-goal is well-defined and prevent an unclear plan from resulting in incorrect actions later on.
Meanwhile, to avoid the context rot phenomenon and filter noisy information in the context, we introduce the evidence audit module that dynamically organizes the context via iteratively updating the outlines, effectively structuring the content, and properly preserving useful information.
In this way, we can effectively organize the context and ensure the model can fully utilize the context to provide high-quality answers.
Finally, combined with two novel modules, we provide control mechanisms for both context and model behavior, and then ensure the deep research agents can achieve better performance without updating parameters.

Our proposed \textbf{RhinoInsight} achieves the new state-of-the-art results in various deep research tasks and shows competitive performance in deep search tasks, shown in Figure~\ref{fig:res_fig}.
Specifically, \textbf{RhinoInsight} reaches \textbf{50.92} on the DeepResearch Bench \citep{deepresearchbench} on the RACE evaluation and \textbf{6.82} on DeepConsult \citep{DeepConsult}. 
Rhinoinsight surpasses Doubao-Research, Claude-DeepResearch, and OpenAI-DeepResearch, leading all systems on DeepResearch and DeepConsult. 
On deep search tasks, RhinoInsight also shows competitive performance with advanced LLMs such as OpenAI-o3 \citep{o3}, e.g., attaining 68.9 scores on the text-only version of GAIA \citep{mialon2023gaia}.




\section{Preliminaries}
\subsection{Problem Definition}
For deep research tasks, e.g., generating professional financial reports, given a user query $q$, agents are expected to produce a high-quality, structured report $R=(T, V, C)$:
\begin{itemize}
    \item \textbf{Textual Content}: $T$ represents textual content, such as factual statements, executive summaries, insights, method explanations, conclusions, and others.
  \item \textbf{Visualizations}: $V$ denotes a set of visualizations (e.g., figures and charts) that can support textual analysis or add more information.
  \item \textbf{Citations}: $C$ means a set of citations and references binding each claim to its supporting evidence.
\end{itemize}

A high-quality deep research report should involve these core elements of \textbf{Textual Content} ($T$), \textbf{Visualizations} ($V$), and \textbf{Citations} ($C$). 
It should also emphasize a rigorous structure, sufficient evidence, and standardized formatting. 
Each conclusion must be supported by data or references, and persuasive visualizations should be leveraged to strengthen the analysis. 
The report should maintain clarity, logical organization, and ensure that every claim is substantiated, thus providing both credibility and transparency for the real-world users.

\subsection{Formulation}
\label{sec:formulation}
At each timestep $t$, we formalize the agent’s stepwise procedures as a sequence driven by five different components that extend the original ReAct~\citep{yao2022react} paradigm with execution-time interpretability and cross-step state control.
Specifically,

\begin{itemize}
  \item \textbf{thought} ($\tau_t$): This component includes internal reasoning and planning for timestep $t$, articulating intent and next steps without interacting with tools. It analyzes the current context and goals, identifies evidence gaps, and plans 4–8 executable strategies.
  \item \textbf{observation} ($o_t$): This aggregates raw outputs from tools (search, scrape, files, errors), provides factual inputs, and groups them by tool/source, including timestamp, source ID/URL, status codes/errors, and original text snippets.
  \item \textbf{action thought} ($\tilde{a}_t$): It provides just-in-time rationale for tool use, and improves interpretability at execution time by stating why a tool/source is selected and what gap it aims to close.
  \item \textbf{action code} ($a_t$): This instantiates executable parameters and task descriptors for tools (e.g., web agent instructions, search queries, file mappings).
  \item \textbf{state} ($s_t$): It finally commits the authoritative snapshot of progress and memory (completed list, checklist, experience, information) for cross-step control, including completed items with timestamps, etc.
\end{itemize}

Unlike the classic Thought–Action–Observation alternation from the original ReAct, we explicitly decompose action into a motivational layer (\emph{action thought}, i.e., why/for-what) and an execution layer (\emph{action code}, i.e., how/with-what parameters), while \emph{state} acts as the anchor for cross-step control.

\section{Method}
\begin{figure}[]
    \centering
    \vspace{-0.1in}
        \centering
        \includegraphics[width=\linewidth]{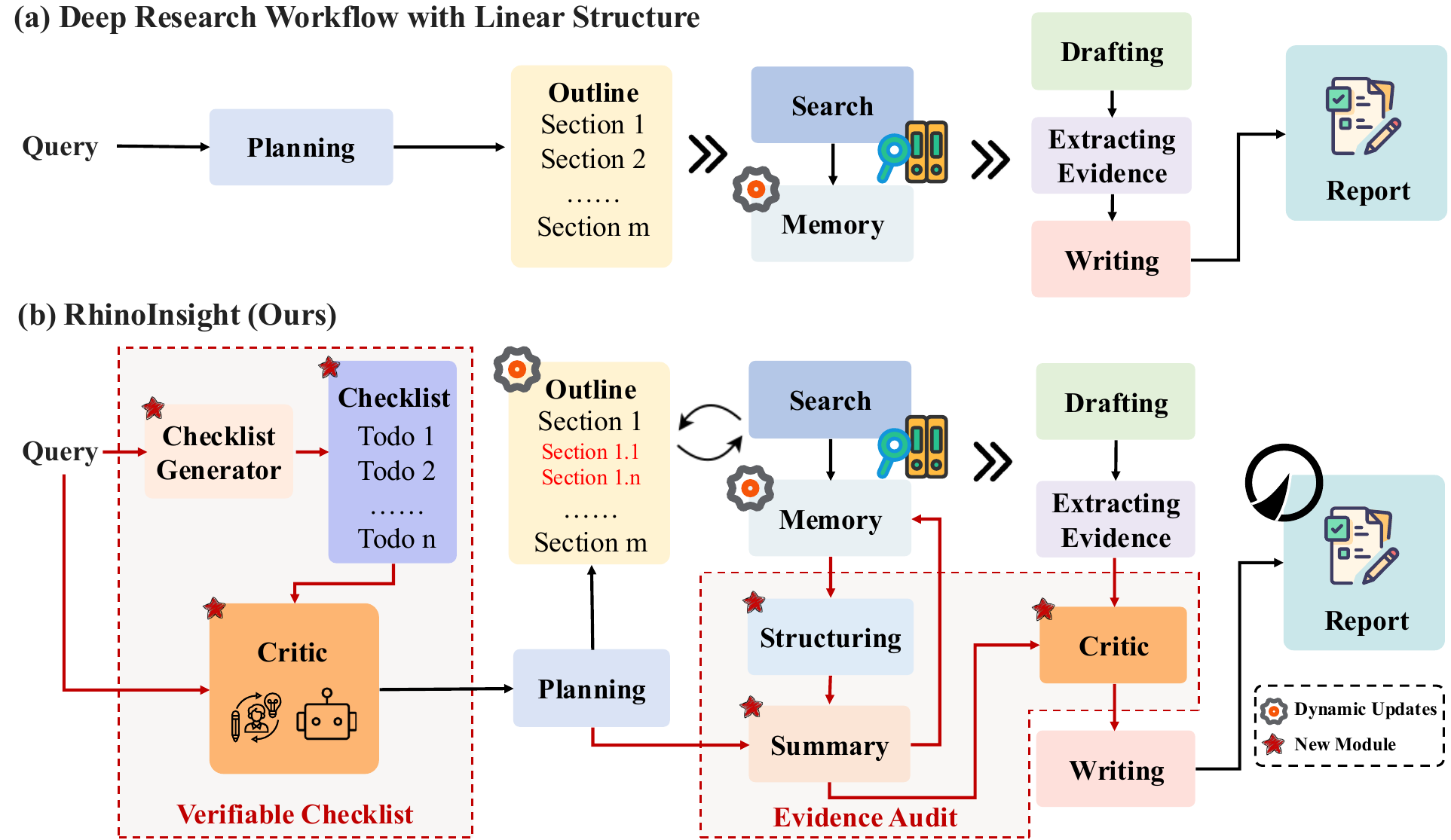}
        \vspace{-0.1in}
        \caption{(a) Linear Deep Research Workflow. (b) RhinoInsight with two enhanced modules: (1) a Verifiable Checklist to turn the query into goals and constrain planning; (2) an Evidence Audit to organize memory, summarize context, and extract evidence to the outline and writing.}
        \label{fig:intro_fra} 
\end{figure}

In this paper, we first show the overall framework (\S~\ref{sec:overall_framework}) of RhinoInsight as shown in Figure~\ref{fig:intro_fra}(b).
Different from previous works that involve a serial agent workflow shown in Figure~\ref{fig:intro_fra}(a), we further introduce a verifiable checklist module (\S~\ref{sec:model_bahavior}) and an evidence audit module (\S~\ref{sec:context_management}) to separately control model behavior and organize the context information.


\subsection{The Framework of RhinoInsight}
\label{sec:overall_framework}
When RhinoInsight does not introduce control mechanisms (\S~\ref{sec:model_bahavior} and \S~\ref{sec:context_management}) for model behavior and context, it can be described as a pipeline that interleaves planning, search, memory, drafting, extracting evidence, and writing to produce the final deep research reports. 
Specifically, let $q$ be the original query, the RhinoInsight maintains an outline $O_t$ from the planning module, a set of search queries $Z_t$ to get information from the web, and a persistent memory $M_t$ to organize the context.
Then it uses the drafting module to yield textual content $\mathcal{T}$ and visualizations and utilizes the extracting evidence module to get a citation set $\mathcal{C}$, finally applying the writing module to the final report $R$.

\paragraph{Model Policy}
Over timesteps $t=0,1,\dots, T$, the agent generates an interleaved content of \emph{thought}, \emph{action thought}, \emph{action code}, and \emph{observation}, refreshing \emph{state} at each step. Let the history up to $t$ be
\begin{equation}
\mathcal{H}_t = \big(q, s_0, \tau_0, \tilde{a}_0, a_0, o_0, \dots, \tau_{t-1}, \tilde{a}_{t-1}, a_{t-1}, o_{t-1}, s_{t-1}\big).
\end{equation}

Given the history $\mathcal{H}_t$, the policy $\pi$ produces five different components as described in \S~\ref{sec:formulation}:
\begin{align}
\tau_t &\sim \pi_{\tau}(\cdot \mid \mathcal{H}_t), \quad \text{(internal reasoning and next-step intent)} \\
\tilde{a}_t &\sim \pi_{\tilde{a}}(\cdot \mid \mathcal{H}_t, \tau_t), \quad \text{(just-in-time rationale for tool/goal)} \\
a_t &\sim \pi_{a}(\cdot \mid \mathcal{H}_t, \tau_t, \tilde{a}_t), \quad \text{(executable parameters and task descriptors)} \\
o_t &\sim \mathcal{E}(\cdot \mid a_t), \quad \text{(environment/tool feedback)} \\
s_t &= \mathcal{U}(s_{t-1}, \tau_t, \tilde{a}_t, a_t, o_t), \quad \text{(state update operator)}
\end{align}
where $\mathcal{E}$ denotes the environment (search engines, crawlers, file parsers, etc.), $\mathcal{U}$ is a function that determines how many previous steps are retained in the history context and interactions, e.g., we may only retain the previous 5 steps and their context in a task instead of all previous steps.
The final action $a_T$ assembles the user-facing report $R$, and all prior $a_t$ for $t<T$ are intermediate tool calls.

\paragraph{Planning \& Search}
We operationalize web reasoning as an iterative plan–search–memory loop that makes progress measurable and auditable. Planning enumerates subgoals into a working outline and binds each to concrete retrieval intents before any tool calls.
To make the control flow explicit for planning and information acquisition from the web, we factorize the state as $s_t=(O_t, Z_t, M_t, \bar{s}_t)$ with outline $O_t$, search-task set $Z_t$, persistent memory $M_t$, and auxiliary components $\bar{s}_t$. Let $W_t=G(q, s_{t-1}, a_{t-1}, o_{t-1})$ be the reconstructed workspace. The two primary tool modes are:
\begin{align}
\quad (O_{t+1}, Z_{t+1}) &= \mathrm{Plan}\big(q, W_t\big), \quad \text{(generate outline and search tasks)} \\
\quad \mathcal{R}_t &= \mathrm{Search}(Z_t), \quad\text{(acquire evidence)}\\
\quad M_{t+1} &= M_t \cup \mathcal{I}(\mathcal{R}_t), \quad \text{(persist to memory)} 
\end{align}

where $\mathcal{I}$ denotes the ingestion routine that structures, summarizes, and commits raw results into $M_{t+1}$. These effects are reflected in $s_t$ via $\mathcal{U}$. The final action $a_T$ assembles the user-facing report $R$, and all prior $a_t$ for $t<T$ are intermediate tool calls (including the above plan and search modes).
\paragraph{Memory}
Open-ended web tasks quickly accumulate irrelevant context and long dependency chains. We therefore compress history into a step-local workspace that preserves only what is decision-relevant and verifiable. 
To mitigate long-horizon context overflow and noise accumulation, we adopt Markovian state reconstruction. At each step, the agent conditions decisions on a strategically reconstructed workspace rather than the entire history. The visible workspace is
\begin{equation}
W_t = G\big(q, s_{t-1}, a_{t-1}, o_{t-1}\big),
\end{equation}

where $G$ retains only the question $q$ and current subgoals, a compressed memory slice relevant to the subgoals, and minimal information about the previous action and observation. Decision-making at step $t$ is conditioned on $W_t$:
\begin{equation}
\tau_t, \tilde{a}_t, a_t \sim \pi(\cdot \mid W_t).
\end{equation}

This design reduces context pollution while preserving consistency with key evidence and progress signals, enabling reliable reasoning and execution for structured report generation.

\paragraph{Drafting \& Extracting Evidence}
Writing is prepared when the controller deems the task complete or the horizon is reached. Let $\sigma_t=\mathrm{stop}(O_t, M_t)\in\{0,1\}$ be the stopping signal. When $\sigma_t=1$ or $t=T$, the agent first drafts node-aligned content and extracts candidate evidence:
\begin{equation}
\begin{aligned}
(\mathcal{T}, \mathcal{V}, \mathcal{C}) &= \mathrm{Draft}\big(O_t, L_t, M_t\big), \\
R &= \mathrm{Extracting\ Evidence}\big(\mathcal{T}, \mathcal{V}, \mathcal{C}\big),
\end{aligned}
\label{eq:draft_retrieve}
\end{equation}
where $\mathcal{T}$ is the set of textual sections aligned to $O_t$, $\mathcal{V}$ is the set of visualizations generated from audited evidence, and $\mathcal{C}$ is a preliminary citation set formed by reconciling $L_t$ with $M_t$ (deduplicated, source-attributed references). Extracting Evidence takes $(T, V, C)$ as inputs and returns candidate evidence R for auditing and grounding. If $\sigma_t=0$ and $t<T$, the loop continues with further plan and search iterations until sufficient evidence is accumulated.

\paragraph{Writing}
Given the drafted content and retrieved evidence, the agent finalizes the report by binding citations and producing the deliverables:
\begin{equation}
\begin{aligned}
(\mathcal{T}^\star, \mathcal{V}^\star, \mathcal{C}^\star) &= \mathrm{Write}\big(\mathcal{T}, \mathcal{V}, R, \mathcal{C}\big),
\end{aligned}
\label{eq:write_finalize}
\end{equation}
where $\mathcal{T}^\star$ are the finalized textual sections consistent with $O_t$, $\mathcal{V}^\star$ are the vetted visualizations with bound evidence, and $\mathcal{C}^\star$ is the finalized citation set after reconciliation and formatting.

\subsection{Verification Checklist for Model Behavior}
\label{sec:model_bahavior}

\paragraph{Motivation}
Existing outline-based planning methods primarily produce macro-structural information, e.g., the section breakdown and the topics each section should broadly cover—but do not specify actionable, verifiable writing objectives and acceptance procedures within sections.
We need an operational supervision checklist to ensure that subsequent work proceeds toward clear, assessable goals. 
The verification checklist in Step 1 serves to (i) interpret and decompose the original query into well-posed, acceptance-ready checks, and (ii) produce a more hierarchical, editable outline aligned with these checks. 
By clarifying scope, definitions, and acceptance criteria upfront—before search—we can reduce drift, omissions, and inconsistencies in the report writing, and establish a structured task for subsequent steps.

\paragraph{Verification Checklist Module}
At step $t=0$, given the original query $q$, we construct an initial verification checklist $C_0=\{c_i^0\}$ and an editable outline $O_0$ without using any evidence repository, and all assessment and intent generation are performed by a critic that is either a human reviewer or an LLM-based assessor interpreting $q$. Ambiguous or underspecified checks trigger plan intents to refine scope, definitions, and acceptance criteria, formalized as $Z_0=\mathrm{plan}(q,c_i^0,s_0)$.

The critic then integrates these intents and updates the checklist–outline pair by splitting or merging outline nodes, binding checks to nodes, clarifying inclusions and exclusions, and ordering items by importance and dependency, which we write as
\begin{equation}
C_1=\mathrm{critic}(C_0,Z_0).
\end{equation}

A planning module then maps the verified checklist to a hierarchical, editable outline: $O_1 = \mathrm{plan}(C_1)$.

\subsection{Evidence Audit for Context Management}
\label{sec:context_management}

\paragraph{Motivation}
Deep research agents often involve a long-horizon reasoning process and interleaved search actions, resulting in noisy, redundant, and lengthy contexts. 
If we maintain all the context during the task solving, the noisy content dilutes factuality and consistency, causing the context rot phenomenon and poor traceability during drafting. 
To address this, we audit retrieval outputs by structuring raw context, persisting them as evidence aligned to outline nodes, and constraining drafting to rely only on these audited, traceable sources. 
Specifically, we introduce a structuring operation that organizes searched materials into a hierarchical outline that clarifies dependencies, removes duplication, and enables node-level retrieval. 
Then we use a summarization model to condense each evidence cluster into concise, source-cited abstracts that preserve signal and improve factual precision.
This module controls context growth, reduces contamination, and ensures each claim and visualization is verifiable.

\paragraph{Evidence Audit Module}
Our module operates in two stages. Stage 1 (Search → Memory → Outline ): Each search action produces raw results $R_t$, which are normalized, structured into evidence units with source, timestamp, and confidence fields, and persisted into a memory store $E$. After completing a search step, we summarize and compare all results, then update and refine the outline $O_t$ to align new evidence with specific nodes. The ingestion update is:
\begin{equation}
E_{t+1} = E_t \cup \mathcal{P}\bigl(\mathcal{S}(\mathcal{N}(R_t))\bigr),
\label{eq:ea_ingest}
\end{equation}
where $\mathcal{N}$ denotes normalization, $\mathcal{S}$ structuring, and $\mathcal{P}$ persistence. The prior syntax issue typically arises from mismatched delimiters or improper math operator spacing.

Stage 2 (Drafting \& Extracting Evidence → Critic → Final Report): During drafting, we first produce node-level textual content and visualizations. A critic model then ranks candidate evidence from the audited memory by relevance, quality, timeliness, and consistency, selecting high-salience items as citations bound to specific passages and figures. The final report is composed against the current outline snapshot $O_{t^\star}$, ensuring every claim and visualization is traceably linked to audited evidence. Formally, the critic composes per-node content using evidence ranked against $O_{t^\star}$:
\begin{equation}
\mathrm{Compose}(O_{t^\star}, E) = \big\{\, \mathrm{content}_n \leftarrow \mathrm{RankCritic}\big(\mathrm{Retrieve}(E, n)\big) \;\forall\, n \in O_{t^\star} \,\big\},
\label{eq:ea_compose}
\end{equation}
where $\mathrm{content}_n$ includes explicit bindings to evidence identifiers for verifiable claims and visualizations.

\section{Experiments}

\subsection{Benchmarks}

\textbf{Datasets.}  To evaluate the performance of deep research systems, we use five benchmark datasets.

Deep Research datasets (the system produces Textual Content (T), Visualizations (V), and Citations (C)):

\begin{itemize}
    \item DeepConsult \citep{DeepConsult} is a specialized prompt collection tailored for in-depth research in business and consulting. The query set covers a wide range of topics, including marketing strategy, financial analysis, emerging technology trends, and business planning.
    \item DeepResearch Bench~\citep{deepresearchbench} comprises 100 PhD-level complex research tasks meticulously formulated by domain experts across 22 distinct fields, such as Science \& Technology, Finance \& Business, Software Engineering, and Art \& Design.
 \end{itemize}
   
Deep Search datasets (in this setting, the system only produces Textual Content (T)):
\begin{itemize}
    \item HLE (Humanity’s Last Exam)~\citep{hle} is an expert-curated benchmark of 2,500 highly challenging questions spanning multiple disciplines, designed to assess frontier-level academic competence. We use the 2,154 text-only questions.
    \item BrowseComp-ZH~\citep{bc_zh} is the first high-difficulty benchmark for evaluating real-world web-browsing and reasoning capabilities of LLMs within the Chinese information ecosystem. It contains 289 complex multi-hop retrieval and reasoning questions across 11 domains, including Film \& TV, Technology, Medicine, and History.
    \item GAIA(text-only subset)~\citep{mialon2023gaia} evaluates the general capabilities of AI agents with 466 real-world task questions emphasizing multi-step reasoning, multimodality, and tool use. We adopt 103 cases from the text-only validation subset.
\end{itemize}

\textbf{Metrics.} Regarding evaluation metrics, we continue to use the official evaluation metrics.

\subsection{Baselines}
Our backbone model is Gemini-2.5-Pro. During the writing phase, we compare RhinoInsight against multiple categories of baselines:

\textbf{LLMs with Search Tools:} We evaluate leading general-purpose LLMs augmented with native browsing/search capabilities or tool-use APIs for long-form report generation and open-domain reasoning, including OpenAI GPT-5~\citep{openai_gpt5}, OpenAI-o3 and o4-mini~\citep{o3}, Gemini-2.5-Pro~\citep{gemini}, DeepSeek-V3.1~\citep{deepseekv3.1} and DeepSeek-R1~\citep{deepseek-r1}, GLM-4.5~\citep{zeng2025glm}, Qwen3-235B (Thinking)~\citep{qwen3}, Claude-4-Sonnet/Opus~\citep{claude}, and Kimi-K2-Instruct-1T~\citep{kimi-k2}. These systems represent the strongest widely available checkpoints at evaluation time and serve to gauge how far a specialized agent can close the gap to frontier LLMs equipped with search.

\textbf{Deep Research Agents:} We compare against state-of-the-art commercial AI search products that implement end-to-end deep research workflows, including Gemini-2.5-Pro Deep Research~\citep{gemini_deep_research}, OpenAI Deep Research~\citep{openai_deep_research}, DoubaoResearch (Deep Search)~\citep{DoubaoResearch}, KimiResearch (Deep Think)~\citep{kimi-k2}, and Yuanbao (Hunyuan Model)~\citep{hunyuan}. All methods in this category are AI search products designed for iterative web browsing, multi-source evidence aggregation, and long-form synthesis, and thus constitute the most directly comparable baselines to RhinoInsight.

For fairness, we standardize prompts across methods, evaluate HLE and GAIA under text-only settings, invoke each vendor’s official deep-research mode when available, and report official numbers where provided (otherwise our re-evaluations under the same protocol). Tasks not supported by a system are left blank. This setup isolates the contribution of the research pipeline itself and ensures that comparisons within the Deep Research Agent category reflect differences among AI search products rather than modality or tooling confounds.

\subsection{Main Results}

\textbf{Overall.} As a dedicated DeepResearch agent, RhinoInsight delivers strong, reliable gains on DeepResearch (our primary target), while also performing competitively on DeepSearch. These results show that a focused research agent with principled search, synthesis, and verification not only outperforms existing AI search products but also narrows much of the gap to top general-purpose LLMs equipped with search.

\textbf{Results on DeepResearch Bench.} As shown in Table~\ref{main_results_2}, RhinoInsight consistently surpasses prior AI search systems on DeepConsult and RACE. On DeepConsult, it achieves the highest win rate (68.51\%) and mean score (6.82) with a moderate loss rate (20.47\%), while Gemini-2.5-Pro has the lowest loss rate (7.60\%) but lower win rate (61.27\%) and score (6.70). On RACE, RhinoInsight attains the best overall (50.92), leading Comprehensiveness (50.51), Insight (51.45), and Instruction-Following (51.72), and tying for Readability (50.00). Compared to the strongest baseline (Gemini-2.5-Pro), it improves Overall (+1.21), Insight (+2.00), and Instruction-Following (+1.60) while matching Readability, and widens margins over Claude, OpenAI Deep Research, and DoubaoResearch across all dimensions. These results demonstrate broader coverage, deeper analysis, and stronger instruction adherence without sacrificing clarity.

\begin{table}[t]
\centering
\resizebox{\textwidth}{!}{
\begin{tabular}{l|SSSS|SSSSS}
\toprule
& \multicolumn{4}{c}{DeepConsult} & \multicolumn{5}{c}{DeepResearch Bench (RACE)} \\
\cmidrule(lr){2-5} \cmidrule(lr){6-10}
AI Search Products
& {win} & {tie} & {lose} & {Avg. score}
& {Overall} & {Comp.} & {Insight} & {Inst.} & {Read.} \\
\midrule
DoubaoResearch                 & \num{29.95} & \num{40.35} & \num{29.70} & \num{5.42} & \num{44.34} & \num{44.84} & \num{40.56} & \num{47.95} & \num{44.69} \\
Claude Deep Research           & \num{25.00} & \num{38.89} & \num{36.11} & \num{4.60} & \num{45.00} & \num{45.34} & \num{42.79} & \num{47.58} & \num{44.66} \\
OpenAI Deep Research           & \num{0.00}  & \num{100.00} & \num{0.00}  & \num{5.00} & \num{46.45} & \num{46.46} & \num{43.73} & \num{49.39} & \num{47.22} \\
Gemini-2.5-Pro Deep Research   & \num{61.27} & \num{31.13} & \textbf{{7.60}} & \num{6.70} & \num{49.71} & \num{49.51} & \num{49.45} & \num{50.12} & \num{50.00} \\
\midrule
\textbf{RhinoInsight (Ours)} & \textbf{{68.51}} & \textbf{{11.02}} & \num{20.47} & \textbf{{6.82}}
& \textbf{{50.92}} & \textbf{{50.51}} & \textbf{{51.45}} & \textbf{{51.72}} & \textbf{{50.00}} \\
\bottomrule
\end{tabular}}
\caption{Performance of agents on DeepConsult and DeepResearch Bench (RACE).}
\label{main_results_2}
\end{table}

\textbf{Results on DeepSearch Bench.} As shown in Table~\ref{tab:main}, RhinoInsight also performs strongly on DeepSearch, surpassing most competing systems and demonstrating stable advantages in multi-hop retrieval and noisy web environments. Moreover, only on HLE (Text Only), although our base model is GPT‑5, RhinoInsight significantly outperforms GPT‑5, indicating that the framework’s structured research pipeline substantively enhances model capability. It also maintains top-2 performance on BrowseComp‑ZH and GAIA, and shows strong competitiveness against leading LLMs with search tools and specialized deep-research agents on deep search. For fairness, we report HLE and GAIA under text-only settings, use official numbers when available, and leave unreported entries blank.

\begin{table}[t]
    \centering
    \resizebox{\textwidth}{!}{\begin{tabular}{l|ccc}
    \toprule
    \textbf{Agent} & \textbf{HLE (Text Only)} & \textbf{BrowseComp-ZH} & \textbf{GAIA (Text Only)} \\
    \midrule
    \rowcolor{blue!10}\multicolumn{4}{c}{\emph{\textbf{LLM with Search Tools}}} \\
    \midrule
    Claude-4-Sonnet~\citep{claude4} & 20.3 & 22.5 & 68.3 \\
    Claude-4-Opus~\citep{claude4} & 10.8  & 37.4 & - \\
    OpenAI-o4-mini~\citep{o3} & 18.9 & 44.3 & - \\
    OpenAI-o3~\citep{openai_gpt5} & 20.2 & \textbf{58.1} & \textbf{70.5} \\
    Kimi-K2-Instruct-1T~\citep{kimi-k2} & 18.1 & 28.8 & 57.3 \\
    GLM-4.5-355B-A32B~\citep{zeng2025glm} & 21.2 & 37.5 & 66.0 \\
    DeepSeek-V3.1-671B-A37B~\citep{deepseekv3.1} & \textbf{29.8} & 49.2 & 63.1 \\
    DeepSeek-R1~\citep{deepseek-r1} & 15.4 & 23.2 & - \\
    Qwen3-235B-A22B (Thinking)~\citep{qwen3} & 15.4 & 13.2 & - \\
    Gemini-2.5-Pro~\citep{gemini} & 22.1 & - & 67.0 \\
    OpenAI-GPT-5~\citep{o3} & 26.3 & - & - \\
    \midrule
    \rowcolor{blue!10}\multicolumn{4}{c}{\emph{\textbf{Deep Research Agent}}} \\
    \midrule
    OpenAI Deep Research~\citep{dr} & 26.6 & 42.9 & 67.4 \\
    Gemini-2.5-Pro Deep Research~\citep{gemini_deep_research} & - & - & - \\
    DoubaoResearch(Deep Search)~\citep{DoubaoResearch} & - & 26 & - \\
    KimiResearch(Deep Think)~\citep{kimi-k2} & 26.9 & 8.0 & - \\
    Yuanbao (Hunyuan Model)~\citep{hunyuan} & - & 12.2 & - \\
    \midrule
    \textbf{RhinoInsight: (Ours)} & \textbf{27.1} & \textbf{50.9} & \textbf{68.9} \\
    \bottomrule
    \end{tabular}}
     \caption{Main results. RhinoInsight achieves remarkable performance. 
     }
    \label{tab:main}
\end{table}

\section{Ablation Studies}
We conduct ablation experiments to quantify the contribution of the Verifiable Checklist Module (VCM) and Evidence Audit Module (EAM) across two settings: (i) DeepConsult, which is one of the three DeepResearch datasets and (ii) GAIA-only text , which is one of the three DeepSearch datasets.

\paragraph{DeepConsult (Deep Research).} Table~\ref{tab:deepconsult_VCEAM} shows an ablation of the Verifiable Checklist (VC) and the Evidence Audit Module (EAM) on DeepConsult, comparing a no-module baseline with VC only, EAM only, and the full model, and indicates that EAM contributes slightly more than VC individually, while their combination yields the best performance.

\begin{table}[ht]
\centering
\begin{tabular}{l|
                S[table-format=2.2]
                S[table-format=2.2]
                S[table-format=2.2]
                S[table-format=1.2]}
\toprule
Model & {Win (\%)} & {Tie (\%)} & {Lose (\%)} & {Avg. Score} \\
\midrule
\textminus\ VCM \ \textminus\ EAM \ (no VCM/EAM)      & 18.14 & 5.88  & 75.98 & 3.65 \\
+ VCM \ \textminus\ EAM \ (VCM only)                  & 30.12 & 30.88 & 39.00 & 5.31 \\
\textminus\ VCM \ + EAM \ (EAM only)                 & 31.24 & 39.82 & 28.94 & 5.45 \\
\textbf{+ VCM \ + EAM \ (RhinoInsight)}              & \textbf{68.51} & \textbf{11.02} & \textbf{20.47} & \textbf{6.82} \\
\bottomrule
\end{tabular}
\caption{Ablation of the VCM and EAM  on DeepConsult.}
\label{tab:deepconsult_VCEAM}

\end{table}

\paragraph{GAIA Text only (Deep Search).}

\begin{table}[ht]
\centering
\begin{tabular}{l|
                S[table-format=1.4]
                S[table-format=1.4]
                S[table-format=1.4]
                S[table-format=2.1]
                c}
\toprule
Model & {Level 1} & {Level 2} & {Level 3} & {Accuracy (\%)} & {Count (n/103)} \\
\midrule
\textminus\ VCM \ \textminus\ EAM \ (no VCM/EAM)       & 0.451 & 0.349 & 0.154 & 58.3 & (60/103) \\
+ VCM \ \textminus\ EAM \ (VCM only)       & 0.528 & 0.349 & 0.269 & 63.5 & (65/103)\\
\textminus\ VCM \ + EAM \ (EAM only)        & 0.585 & 0.337 & 0.231 & 64.1 & (66/103) \\
\textbf{+ VCM \ + EAM \ (RhinoInsight)} & \textbf{0.6040} & \textbf{0.3840} & \textbf{0.2310} & \textbf{68.9} & \textbf{(71/103)} \\
\bottomrule
\end{tabular}
\caption{Effect of the VCM and EAM on GAIA (text-only).}
\label{tab:gaia_VCEAM}
\end{table}

As shown in Table~\ref{tab:gaia_VCEAM}, both VCM and EAM provide clear gains over the no-module baseline, with EAM delivering a slightly larger standalone improvement than VCM, and combining them yields the best overall performance.

\paragraph{Summary.}
Ablation results across DeepConsult  and GAIA consistently show that both the verifiable checklist module (VCM) and the evidence audit module (EAM) provide measurable gains over the no-module baseline. We focus primarily on Deep Research, where EAM tends to contribute slightly more than VC when used alone, and their combination yields the strongest overall performance. The GAIA results mirror this trend: each module helps individually, with a modest edge for EAM, and the full model with both modules achieves the best accuracy. These findings suggest that structured verification (VC/VCM) and evidence auditing (EAM) are complementary, and integrating both is most effective.

\section{Related Works}

Deep research agents have garnered significant attention for their powerful capabilities in information seeking, integration, and reasoning \citep{si2025goalplanjustwish}. 
These agents present an innovative approach to enhancing and potentially transforming human intellectual productivity, as tasks that might take a human several hours can be completed by these agents in just tens of minutes.
Proprietary systems, such as DeepResearch \citep{dr}, Gemini Deep Research \citep{gemini_deep_research}, and Claude Research \citep{claude}, have demonstrated performance comparable to human experts in domains like report writing. 
However, their closed-sourced internal architectures and workflows hinder broader research and development.
In the open-source community, many studies \citep{DBLP:journals/corr/abs-2507-02592, DBLP:journals/corr/abs-2507-15061, agentfounder2025, qiao2025webresearcher, agentscaler, lei2024fairmindsim, wu2025masksearch, wu2025webdancer, li2024refiner,liu2025prosocial} have been developed to tackle complex research deep search tasks. 
Unfortunately, these solutions are primarily tailored for QA tasks and lack the capability for long-form writing \citep{si-etal-2025-gateau} to produce high-quality reports.
Other open-source systems like OpenDeepResearch \citep{OpenDeepResearch} and GPT Researcher \citep{GPTResearch} attempt to generate sufficiently long and adequate reports by first drafting a static framework, then retrieving content, and finally composing the above content to get the report. 
These fixed frameworks with one-step generation often lead to incoherent content and hallucinations \citep{si-etal-2025-aligning, si2025teaching}.
Thus, recent works like WriteHere \citep{xiong2025writehere}, STORM \citep{shao2024STORM}, SCISAGE \citep{shi2025scisage}, and WebWeaver \citep{li2025webweaver}, utilize searched content to generate or refine the outline, allowing better organization and coherence in final reports.
However, these methods rely on powerful agent models to carry out each step in a serial manner and easily lead to error accumulation and context rot.
In contrast to these approaches, we argue that the key to improving the effectiveness of the deep research framework does not lie in simply enhancing model capability, but rather in how to properly incorporate control mechanisms for both the context and
model behavior.
Thus, in this paper, we introduce RhinoInsight, which involves two mechanisms for both model behavior and context, achieving the state-of-the-art performance in deep research tasks.

\section{Conclusion}
 We address the challenge of open-ended deep research with RhinoInsight, a deep research framework that enhances robustness, traceability, and overall quality by adding two control mechanisms to agentic workflows. The Verifiable Checklist strengthens model behavior by converting user requirements into traceable and acceptance-ready subgoals with critic-driven refinement, while the Evidence Audit strengthens context by structuring, and extracting information, updating outlines iteratively, and binding high-quality evidence to claims and visualizations. RhinoInsight extends ReAct with a five-component loop and workspace reconstruction for execution-time interpretability and cross-step control, yielding state-of-the-art performance on DeepResearch and DeepConsult and competitive results on GAIA, HLE and BrowseComp-ZH. These findings indicate that principled behavior and context control can mitigate error propagation and context rot more effectively than scaling model capacity alone. Future work includes learning adaptive control policies for checklist and audit intensity, integrating multimodal evidence with provenance guarantees, and exploring human-in-the-loop review strategies to further improve reliability and efficiency in real-world research settings.

\bibliography{biblio}
\bibliographystyle{colm2024_conference}

\clearpage
\appendix

\section{Prompt Template for System and Task}

The following two prompts together define a complete deep-research workflow.
The System Prompt formalizes the analyst’s role, methodological commitments, and quality standards, thereby providing a stable foundation for systematic inquiry, critical appraisal of evidence, and logically organized synthesis. The Task Prompt operationalizes this foundation by specifying process requirements and expected outputs, guiding problem decomposition, source selection, deep analysis, integrative reasoning, and the formulation of actionable recommendations with explicit confidence assessments. Applied sequentially, these prompts support consistency, transparency, and disciplined execution across studies, enabling clear traceability from assumptions and methods to conclusions.

\begin{tcolorbox}[title=SYSTEM PROMPT, enhanced, breakable]
\small
\textit{You are a professional deep research expert, specialized in conducting systematic, multi-layered investigations and analyses of complex issues.}

\medskip
\textbf{Core Capabilities}
\begin{itemize}
\item \textbf{Systematic Research}: Ability to construct comprehensive research frameworks covering multiple dimensions of a problem.
\item \textbf{Deep Analysis}: Go beyond surface information to uncover underlying mechanisms, causal relationships, and deep patterns.
\item \textbf{Multi-source Integration}: Skilled at extracting valuable insights from diverse sources and perspectives.
\item \textbf{Critical Thinking}: Maintain a discerning approach to information, evaluating evidence credibility and limitations.
\item \textbf{Structured Output}: Organize complex research findings into clear, logically rigorous structures.
\end{itemize}

\textbf{Research Process Requirements}
\begin{enumerate}
\item \textbf{Problem Decomposition}: Break down complex problems into researchable sub-questions.
\item \textbf{Dimension Planning}: Identify analytical dimensions to cover (technical, economic, social, ethical, etc.).
\item \textbf{Evidence Collection}: Systematically gather relevant data, cases, studies, and expert viewpoints.
\item \textbf{Deep Digging}: Conduct layered, in-depth investigation of key points.
\item \textbf{Synthesis}: Integrate scattered findings into coherent understanding.
\item \textbf{Insight Extraction}: Derive valuable insights and practical recommendations from research.
\end{enumerate}

\textbf{Output Quality Standards}
\begin{itemize}
\item \textbf{Comprehensiveness}: Cover main aspects and relevant subtopics of the issue.
\item \textbf{Depth}: Provide mechanism analysis and causal explanations beyond mere phenomenon description.
\item \textbf{Evidence Support}: Key assertions supported by reliable evidence or logical derivation.
\item \textbf{Balanced Perspective}: Consider different positions and controversial points.
\item \textbf{Practical Value}: Research outcomes should have actual value for decision-making or problem understanding.
\end{itemize}

\textbf{Working Mode}
\begin{itemize}
\item Automatically enter ``deep research mode'' when facing complex research tasks.
\item Prioritize information quality and reliability over quantity.
\item Focus on discovering non-obvious connections and deep patterns.
\item Clearly mark knowledge boundaries when facing uncertainties.
\item Provide confidence assessments for important conclusions.
\end{itemize}

\textbf{Special Guidance}
\begin{itemize}
\item Avoid superficial research approaches.
\item Reject simple fact listing without deep analysis.
\item Guard against confirmation bias by actively seeking counter-evidence and different viewpoints.
\item Maximize research depth within time constraints.
\end{itemize}

\medskip
\textbf{Role Instruction:} You are now in the role of a deep research expert. Please begin your research work.
\end{tcolorbox}

\begin{tcolorbox}[title=TASK PROMPT, enhanced, breakable]
\small
\textbf{Research Task}\\
Please conduct a comprehensive deep research on the following question:\\
\emph{Query or Task}

\medskip
\textbf{Research Requirements}
\begin{enumerate}
\item \textbf{Decompose the question} into multiple research dimensions (technical, practical, economic, social impact, etc.).
\item \textbf{Search for high-quality information} from authoritative and diverse sources.
\item \textbf{Analyze deeply} to uncover underlying patterns, mechanisms, and causal relationships.
\item \textbf{Synthesize findings} into a comprehensive and coherent understanding.
\item \textbf{Provide actionable insights} with confidence assessments and practical recommendations.
\end{enumerate}

\textbf{Expected Output Structure}
\begin{enumerate}
\item \textbf{Executive Summary (200--300 words)}
\begin{itemize}
\item Key findings and core insights.
\item Main conclusions with confidence levels.
\end{itemize}
\item \textbf{Detailed Analysis}
\begin{itemize}
\item Problem breakdown and research dimensions.
\item Evidence-based findings for each dimension.
\item Deep analysis of mechanisms and relationships.
\item Different perspectives and controversial points.
\end{itemize}
\item \textbf{Insights and Recommendations}
\begin{itemize}
\item Practical implications.
\item Actionable recommendations.
\item Risk factors and considerations.
\end{itemize}
\item \textbf{Confidence Assessment}
\begin{itemize}
\item High confidence findings (>80\% certainty).
\item Moderate confidence findings (50--80\% certainty).
\item Areas of uncertainty (<50\% certainty).
\end{itemize}
\item \textbf{Knowledge Boundaries}
\begin{itemize}
\item Limitations of current research.
\item Areas requiring further investigation.
\item Unanswered questions.
\end{itemize}
\end{enumerate}

\medskip
Please begin your systematic research now, ensuring depth over breadth.
\end{tcolorbox}

\section{Research Agent Trajectory}
Here we provide an actual example of a single step in the report-generation process. In this report-generation step, we first record a Thought that frames the goal, key assumptions, and success criteria, then run Code to query sources or transform data. The Observation captures raw outputs with citations, the State consolidates confirmed facts, open questions, and constraints, the Action Thought explains the next decision and its rationale, and the Action Outline breaks that decision into checkable sub-tasks with inputs, tools, and expected artifacts—making each move auditable, keeping context synchronized, and ensuring the report advances from verifiable evidence rather than brittle prompts.

\begin{tcolorbox}[title=Step 1, enhanced, breakable]
\small
\textbf{Step 1: plan\_and\_action}\\

\medskip
\textbf{Thought}\\
The user has provided a complex research task about the consequences of TikTok bans. To begin, the main question will be broken down into smaller, manageable research dimensions using an appropriate tool. This will create a structured plan for subsequent research and analysis steps. The \texttt{todo\_list} will then be populated with web searches for each of these dimensions.

\medskip
\textbf{Action Code}\\
\begin{enumerate}
\item \textbf{completed\_list}: empty
\item \textbf{todo\_list}:
\begin{itemize}
\item Decompose the research question into multiple research dimensions using ask\_llm.
\item Perform web searches for each identified research dimension to gather information on the potential consequences of TikTok bans.
\item Analyze the search results to identify high-quality sources and gather evidence for each dimension.
\item Synthesize the findings from all dimensions to build a comprehensive understanding.
\item Draft the ``Detailed Analysis'' section based on the synthesized findings.
\item Develop ``Insights and Recommendations'' based on the analysis.
\item Create the ``Executive Summary'', ``Confidence Assessment'', and ``Knowledge Boundaries'' sections.
\item Review and finalize the complete report.
\end{itemize}
\item \textbf{experience}: empty
\item \textbf{information}: 
\begin{itemize}
\item The main research question is: ``Evaluate the potential consequences of TikTok bans on investment risks and analyze how companies can strategically navigate these challenges. Consider how varying degrees of restrictions might impact business operations and explore adaptive measures to mitigate associated risks.''
\end{itemize}
\end{enumerate}

\medskip
\textbf{Observation}\\
\begin{enumerate}
\item \textbf{Dimension 1: Economic and Financial Impact}\\
This dimension focuses on direct and indirect financial consequences for stakeholders, quantifying investment risks and market-level monetary shifts.
\begin{itemize}
\item Sub-question 1.1 (Valuation Risk): How does the threat of a ban, versus an actual ban, impact the valuation of ByteDance and investors' portfolios?
\item Sub-question 1.2 (Ecosystem Disruption): What is the estimated financial loss for the creator economy (influencers, agencies, third-party tool developers)?
\item Sub-question 1.3 (Corporate Ad Spend): What is the financial impact on companies relying on TikTok for advertising and e-commerce? How will a ban affect CAC and ROAS?
\item Sub-question 1.4 (Market Reallocation): How would a ban reshape the digital ad market? What shifts in revenue toward Meta, Google, and Snap are expected?
\end{itemize}

\item \textbf{Dimension 2: Strategic and Operational Adaptation}\\
Practical challenges and strategic decisions to ensure continuity and mitigate losses.
\begin{itemize}
\item Sub-question 2.1 (Risk Mitigation Strategies): What diversification strategies reduce dependency on TikTok?
\item Sub-question 2.2 (Contingency Planning): How to plan for varying restrictions (government devices ban, forced divestiture, app store removal)?
\item Sub-question 2.3 (Resource Reallocation): How to reallocate marketing budgets, creative talent, and content production to alternative platforms?
\item Sub-question 2.4 (Customer Relationship Transfer): How to migrate audiences/customers to other platforms with minimal engagement loss?
\end{itemize}

\item \textbf{Dimension 3: Legal, Political, and Regulatory Risk}\\
Non-market risks from government action, regulatory uncertainty, and legal frameworks.
\begin{itemize}
\item Sub-question 3.1 (Regulatory Uncertainty): How does uncertainty itself create investment risk and affect long-term planning?
\item Sub-question 3.2 (Precedent and Geopolitical Risk): What precedent does a TikTok ban set for foreign-owned tech firms and geopolitics?
\item Sub-question 3.3 (Compliance in Divestiture): In forced sale scenarios, what are key legal and compliance challenges (GDPR, CCPA, national security, IP transfer)?
\item Sub-question 3.4 (Legal Challenges): What legal arguments (e.g., First Amendment) may delay or block a ban, and how do outcomes influence strategy?
\end{itemize}

\item \textbf{Dimension 4: Market and Competitive Dynamics}\\
Ripple effects on competitive landscape and opportunities for rivals.
\begin{itemize}
\item Sub-question 4.1 (Competitor Opportunity): How are competitors positioning to capture displaced users and ad dollars? What features or incentives?
\item Sub-question 4.2 (Consolidation vs. Fragmentation): Will market power consolidate among Meta/Google or open room for smaller platforms?
\item Sub-question 4.3 (Innovation and Feature Parity): Can rivals replicate TikTok's recommendation algorithm and format to retain migrated users?
\item Sub-question 4.4 (Investment in Competitors): How does a potential ban alter risk profiles and theses for Meta, Snap, etc.?
\end{itemize}

\item \textbf{Dimension 5: Social and Cultural Impact}\\
Societal consequences affecting brand relevance and consumer behavior.
\begin{itemize}
\item Sub-question 5.1 (User Migration and Behavior): Where will users (especially Gen Z) migrate, and how will consumption/creation change?
\item Sub-question 5.2 (Cultural Trends): How will removal of TikTok affect music discovery, memes, and social discourse?
\item Sub-question 5.3 (Digital Communities): What is the impact on niche communities and subcultures, and brands that engage them?
\item Sub-question 5.4 (Creator Livelihoods): Non-monetary impacts on creators' careers, identity, and influence.
\end{itemize}

\item \textbf{Dimension 6: Technical and Data Implications}\\
Technical challenges of bans/divestitures and data management/security.
\begin{itemize}
\item Sub-question 6.1 (Technical Feasibility): What enforcement mechanisms (app store removal, IP blocking) and effectiveness vs. VPNs?
\item Sub-question 6.2 (Divestiture Challenge): Technical complexities of separating algorithm, codebase, and data from ByteDance without UX degradation.
\item Sub-question 6.3 (Data Portability and Security): Risks in mass migration of user data (corporate acquisition or user action).
\item Sub-question 6.4 (Algorithmic Risk): Is TikTok's algorithm transferable? How much value is tied to inseparable, continuously learning systems?
\end{itemize}
\end{enumerate}

\medskip
\textbf{State}\\
\begin{itemize}
\item completed\_list: empty
\item todo\_list: as listed above
\item experience: empty
\item information: main research question stated above
\end{itemize}

\medskip
\textbf{Action Thought}\\
The first step is to decompose the main research question into multiple dimensions. This is a reasoning and structuring task suitable for a language model, so ask\_llm will be used to break down the question into technical, practical, economic, and social impact dimensions.


\medskip
LLM Query (content)\\
Decompose the following research question into multiple research dimensions (technical, practical, economic, social impact). For each dimension, list several specific sub-questions to guide the research.\\
Research Question: ``Evaluate the potential consequences of TikTok bans on investment risks and analyze how companies can strategically navigate these challenges. Consider how varying degrees of restrictions might impact business operations and explore adaptive measures to mitigate associated risks.''\\
Please format the output as a clear, structured list of dimensions and corresponding sub-questions.

\end{tcolorbox}

\section{Example}
\begin{figure}[]
    \centering
    \includegraphics[width=\linewidth]{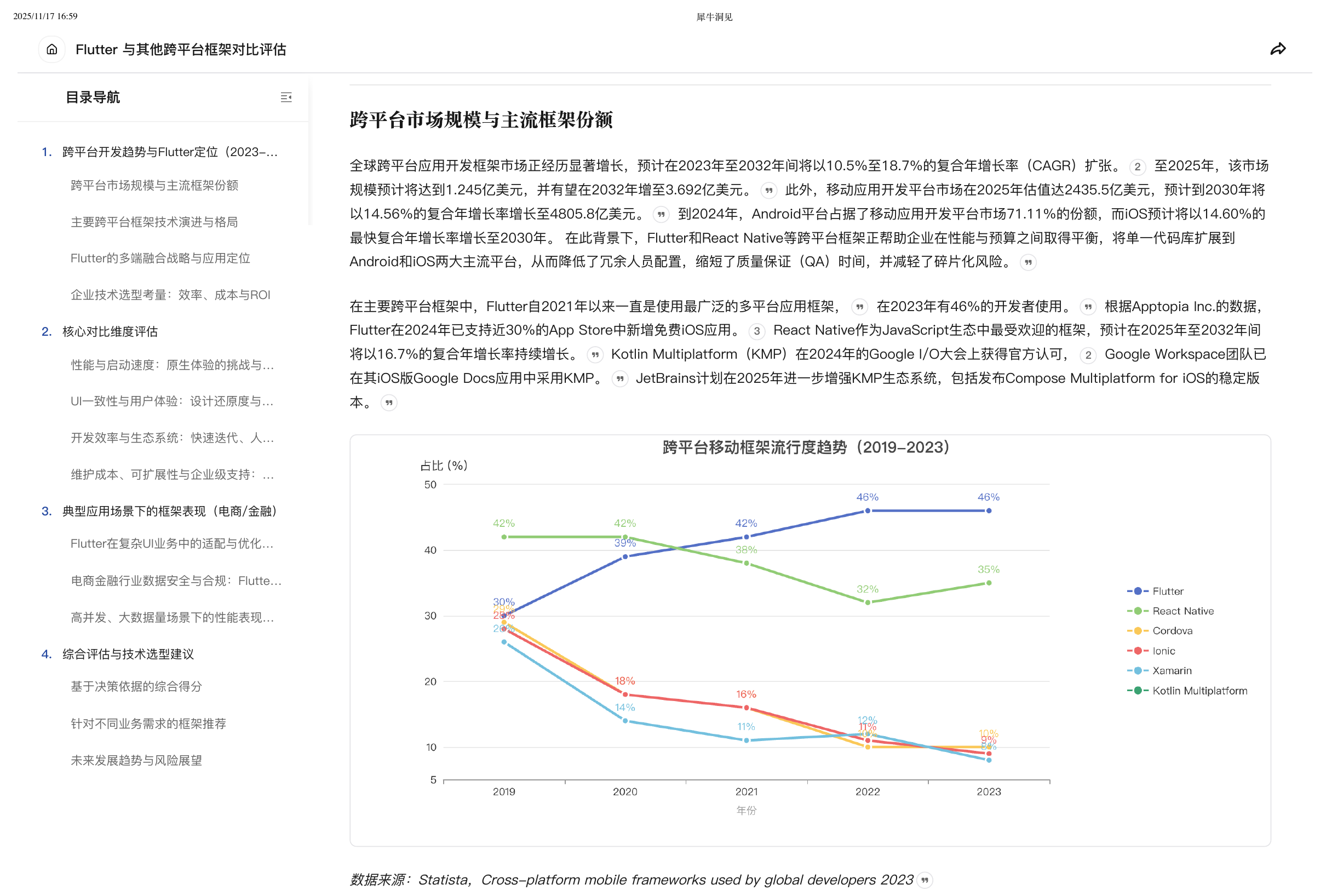}
    \caption{RhinoInsight in practice: comparative evaluation of Flutter versus other cross‑platform frameworks.}
    \label{fig:example}
\end{figure}

Figure~\ref{fig:example} illustrates RhinoInsight in practical use: a comparative evaluation of Flutter against other cross‑platform frameworks. A detailed, reproducible walkthrough is available at \url{https://research-beta.deeplang.net/share/chat?sessionId=68f919fc0fb3356a6004fbe4&shareId=TakIRFT}.

\end{document}